%% file: root.tex
\documentclass[letterpaper, 10 pt, conference]{ieeeconf}  

\IEEEoverridecommandlockouts                              

\usepackage{graphics} 
\usepackage{epsfig} 
\usepackage{mathptmx} 
\usepackage{times} 
\usepackage{amsmath} 
\usepackage{amssymb}  
\usepackage{marvosym}
\usepackage{mathtools}
\let\labelindent\relax
\usepackage{enumitem}
\usepackage{booktabs}
\usepackage{makecell}
\usepackage{xcolor}
\usepackage{tabularx}
\usepackage{hyperref}

\title{\LARGE \bf
 Closed-Loop Action Chunks with Dynamic Corrections for Training-Free Diffusion Policy
}

\author{Pengyuan Wu$^{1,2*}$, Pingrui Zhang$^{2,3*}$,
Zhigang Wang$^{2}$, Dong Wang$^{2}$, Bin Zhao$^{2,4\textsuperscript{\Letter}}$, Xuelong Li$^{5}$, \textit{Fellow,IEEE }
\thanks{$^{1}$Zhejiang University. $^{2}$Shanghai Artificial Intelligence Laboratory. $^{3}$Fudan University. $^{4}$Northwestern Polytechnical University. $^{5}$Institute of Artificial Intelligence, China Telecom Corp Ltd.}%
\thanks{Emails: {\tt\small \{pengyuanwu0810, zprzpr121\}@gmail.com}}
\thanks{$*$ Equal contribution. \textsuperscript{\Letter} Corresponding author.}%
}

\begin{document}

\maketitle
\thispagestyle{empty}
\pagestyle{empty}

\begin{abstract}
Diffusion-based policies have achieved remarkable results in robotic manipulation but often struggle to adapt rapidly in dynamic scenarios, leading to delayed responses or task failures. We present DCDP, a Dynamic Closed-Loop Diffusion Policy framework that integrates chunk-based action generation with real-time correction. DCDP integrates a self-supervised dynamic feature encoder, cross-attention fusion, and an asymmetric action encoder-decoder to inject environmental dynamics before action execution, achieving real-time closed-loop action correction and enhancing the system's adaptability in dynamic scenarios. In dynamic PushT simulations, DCDP improves adaptability by 19\% without retraining while requiring only 5\% additional computation. Its modular design enables plug-and-play integration, achieving both temporal coherence and real-time responsiveness in dynamic robotic scenarios, including real-world manipulation tasks. The project page is at: \url{https://github.com/wupengyuan/dcdp}
\end{abstract}

\input{section/1_introduction}
\input{section/2_related_work}

\input{section/3_method}
\input{section/4_experiment}
\input{section/5_conclusion}

\section*{ACKNOWLEDGMENT}
This work was supported by the Shanghai AI Laboratory, the National Key Research and Development Project (2024YFC3015503), the National Natural Science Foundation of China (62376222), and the Natural Science Basic Research Program of Shaanxi (2025JC-TBZC-07).

\bibliographystyle{IEEEtran}
\bibliography{IEEEabrv,reference}
\end{document}

%% file: section/1_introduction.tex
\section{Introduction}
\label{sec:intro}
In recent years, diffusion policies have achieved remarkable results in robotic manipulation tasks~\cite{chiDiffusionPolicyVisuomotor2023,reuss2023goal,liu2025immimiccrossdomainimitationhuman}. These methods typically reason over action chunks to capture non-Markovian dependencies, reducing compounding errors in sequential prediction and enabling coherent long-horizon action generation~\cite{janner2022planning,pearce2023imitating}.

\begin{figure}[tbp]
\centering
\includegraphics[width=0.99\linewidth]{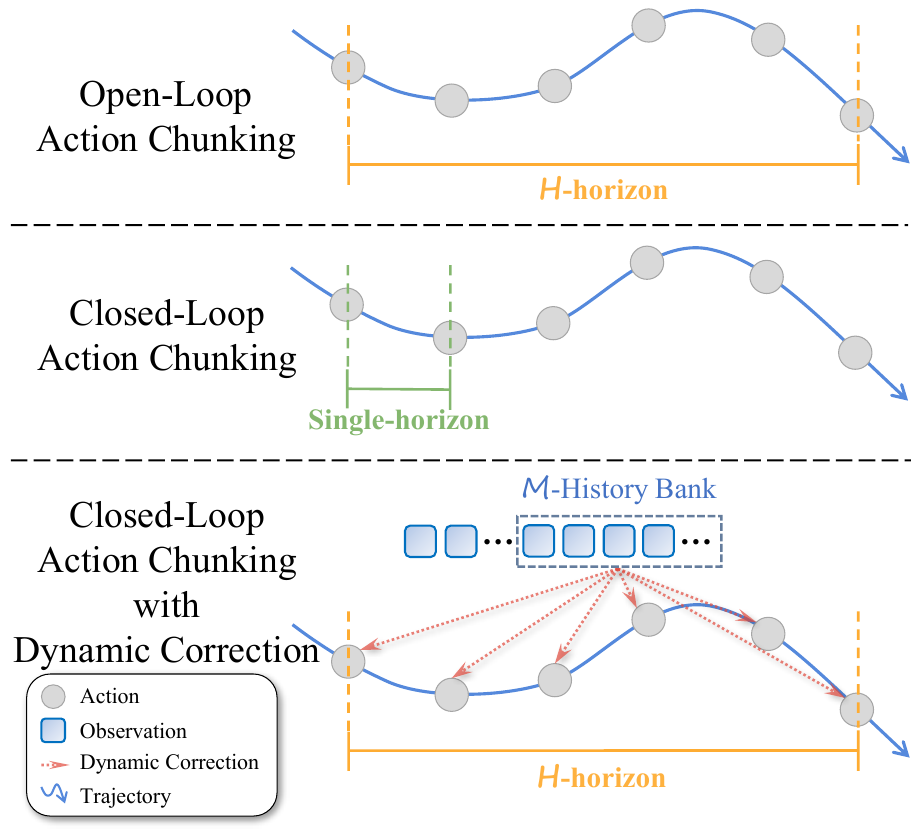}
\caption{\textbf{Comparison of the Open-Loop Diffusion Policy, Closed-Loop Diffusion Policy, and Our Closed-Loop with Dynamic Correction Diffusion Policy.} 
The \textcolor{orange}{orange line} depicts the action chunking prediction of length $H$ for the Open-Loop Diffusion Policy.
Meanwhile, the \textcolor{green}{green line} denotes the Closed-Loop Diffusion Policy, which employs single-step inference to achieve closed-loop control. However, this approach incurs high latency and requires frequent re-planning. By contrast, our Dynamic Correction Closed-Loop Diffusion Policy (as indicated by the \textcolor{red}{red line}) leverages information from a length-$M$ History Bank to perform lightweight, fast corrections at each inference step, thereby achieving closed-loop control. 
}
\vspace{-10pt}
\label{fig:compare}
\end{figure}

However, achieving efficient robotic manipulation in dynamic environments requires more than long-term planning; the policy must also be capable of promptly responding to rapid environmental changes. 
This dual demand poses a significant challenge to existing approaches: the policy must generate coherent action sequences over extended time horizons while simultaneously perceiving and adapting to external disturbances or target motions during execution.

Without such responsiveness, open-loop execution often leads to delays or task failures. 
In typical scenarios, such as grasping moving objects, the lack of sufficient reactivity can drastically reduce success rates.

Generally, the limitations of existing approaches can be summarized as follows:
(1) Open-loop action generation. Action chunks are fully generated prior to execution, lacking the capability to dynamically adjust subsequent actions based on the latest observations. This severely restricts responsiveness in dynamic environments.
(2) Insufficient temporal modeling. The inference process often relies on single-frame or a small number of static observations, failing to fully exploit continuous temporal information, which in turn limits adaptability to environmental changes.

To address the above issues, previous studies have explored two primary avenues for improvement. The first increases inference frequency by shortening the prediction horizon or reducing denoising steps to enhance system responsiveness~\cite{prasad2024consistency,chiDiffusionPolicyVisuomotor2023,janner2022planning}. However, these approaches generally compromise action generation quality, and excessively frequent updates may disrupt action sequence continuity, thereby weakening task execution stability. The second avenue employs temporal ensemble or bidirectional decoding to achieve closed-loop execution by integrating multi-step inferred action sequences~\cite{liu2025bidirectional,george2023one}. Nevertheless, these methods face two main limitations: they are either constrained by historical inference information, hindering true real-time closed-loop execution, or they require sampling multiple action sequences within a single inference step for decoding and selection, which incurs substantial computational overhead.

Facing these challenges, this paper proposes a closed-loop action-chunking policy framework that integrates long-term planning with real-time dynamic correction. The framework leverages the strengths of diffusion policies in long-horizon planning, while incorporating a lightweight dynamic feature module that injects high-frequency environmental information into action generation. In this way, closed-loop execution and dynamic responsiveness are achieved at every time step.

Specifically, we first design a self-supervised dynamic feature encoder to extract information about environmental changes. This encoder collects recent observation images within a sliding window and applies self-supervised contrastive learning on their differential features, thereby capturing high-frequency dynamics. To further enhance temporal awareness, we introduce cross-attention and temporal-attention modules, which strengthen feature modeling in dynamic scenes. In addition, we develop an asymmetric action encoder that compresses raw action sequences into latent representations and reconstructs them using dynamic feature information. By enforcing reconstruction loss together with a KL divergence constraint, the decoder is compelled to rely on recent dynamic observations, thereby improving the adaptability of action generation.

Generally, the training and inference stages of the proposed method can be illustrated as follows: 

Stage 1 (Training):Using labeled data, the asymmetric action encoder and the self-supervised dynamic feature encoder are trained end-to-end. Reconstruction loss and KL divergence are employed to guide the decoder to attend to dynamic features.

Stage 2 (Inference):The pretrained diffusion policy generates complete action chunks to ensure long-term action consistency. Simultaneously, the dynamic feature module continuously extracts environmental change information through a sliding window and jointly decodes it with the action latent representations, enabling real-time correction of actions at each time step. Because the module updates at the same frequency as the action execution, it can significantly enhance adaptability to dynamic environments while maintaining overall action coherence.

It is worth emphasizing that this approach does not require any retraining of the original diffusion policy. By simply inserting the dynamic correction module during inference, the framework substantially improves responsiveness and robustness in dynamic scenarios. Moreover, the framework is highly modular and compatible, allowing seamless integration with various action-chunk-based policies and providing a plug-and-play solution to balance long-term planning with real-time closed-loop control.

We evaluated the DCDP method on the dynamic PushT simulation task and two real-world tasks. The results demonstrate that DCDP can effectively mitigate the adaptability limitations of Diffusion Policy in dynamic scenarios without requiring retraining. Compared with the original inference method, it achieved a 19\% improvement in success rate with only ~5\% additional computational overhead. Furthermore, in static scenarios, its efficient closed-loop characteristic further improved task success.

In summary, the contributions of this work are as follows:

\begin{itemize}[left=0pt]
  \item \textbf{Dynamic Closed-Loop Framework}: Integrates long-horizon planning with real-time correction, preserving Diffusion Policy consistency while enabling fast responses to environmental changes.
  \item \textbf{Dynamic Feature Extraction and Action Correction}: Lightweight temporal attention module learns environmental dynamics and fuses them with latent actions, allowing flexible adaptation to perturbations and moving targets.
  \item \textbf{Training-Free and Modular Design}: Enhances dynamic adaptability without retraining Diffusion Policy; modular design supports plug-and-play integration with various action-chunk strategies for long-term planning and real-time control.
\end{itemize}

%% file: section/2_related_work.tex
\section{Related Work}
\label{sec:rw}

\subsection{Behavior Cloning}
\label{sec:BC}
Recent advancements in data collection and benchmark development, both in simulated and real-world environments~\cite{vuong2023open,khazatsky2024droid,walke2023bridgedata,Zhang_2025_ICCV}, have significantly contributed to the progress of robotics. 
Imitation learning, particularly through expert demonstrations, has emerged as a pivotal driver in advancing robotic capabilities~\cite{torabi2018behavioral,kim2024openvla,liu2024rdt,black2410pi0,liang2025mm,yang2025instructvla,chen2025internvla,zhang2025crossleftrightbrain}. 
Among the various techniques in imitation learning, Generative Behavior Cloning has garnered significant attention due to its ability to effectively model the distribution of expert demonstrations.  
This approach not only simplifies the learning process but has also shown strong empirical success in real-world applications~\cite{brohan2022rt,chi2024universal,zhao2023learning}. 
Recently, a Behavior Cloning method incorporating action chunking has been proposed~\cite{chiDiffusionPolicyVisuomotor2023,swamy2022causal,george2023one}. This technique effectively manages temporal dependencies by predicting sequences of continuous actions. 
However, the process of inferring action chunks often depends on single-frame or a limited number of frame observations, which fails to fully leverage the continuous temporal information present in the data. 
In contrast, our approach exploits previously underutilized temporal cues in the observations and integrates a fast policy to inject these signals into a slower diffusion policy, facilitating online policy correction. 

\subsection{Closed-Loop Action Chunks}

The diffusion policy generates high-quality action through iterative refinement from Gaussian noise, resulting in significant improvements in robotic manipulation~\cite{chiDiffusionPolicyVisuomotor2023,reuss2023goal,pearce2023imitating}.
However, the re-planning frequency of the diffusion policy is limited, and the execution of an action chunking process remains open-loop.
The poor performance of the diffusion policy in rapidly changing environments is also attributed to the open-loop nature of action chunking, which prevents it from receiving timely feedback and responding dynamically.
To address this, some approaches~\cite{xue2025reactive,ye2025ra,shi2024yell} aim to incorporate high-frequency policies, injecting additional information during the execution of an action chunk to enable closed-loop control.
We also adopt a closed-loop scheme, enhancing the dynamic capabilities of the diffusion policy by fully utilizing the memory and differential dynamic information within the closed-loop action chunks.

\begin{figure*}[htbp]
\centering
\includegraphics[width=0.99\linewidth]{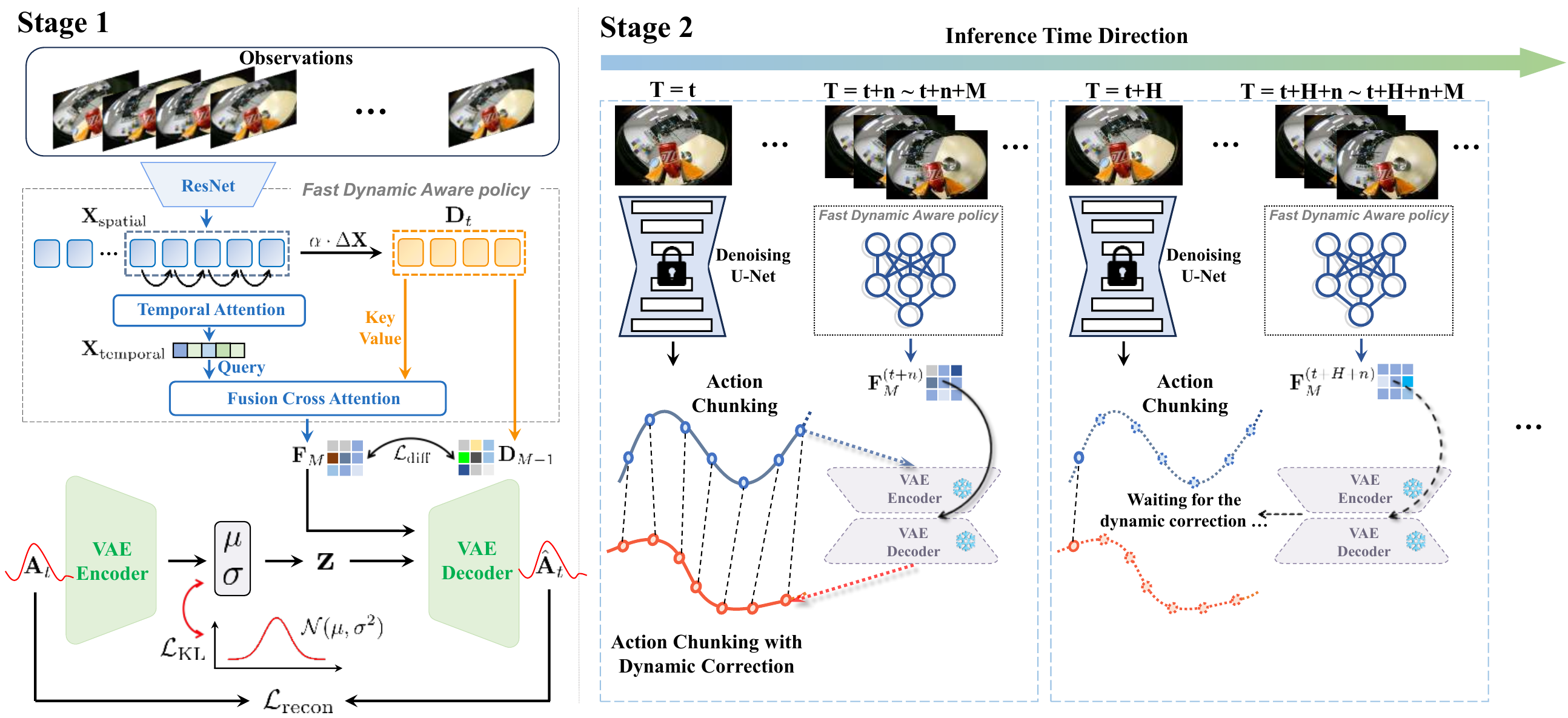}
\vspace{-10pt}
\caption{\textbf{Overview of the DCDP.} Our method adopts a two-stage framework. In \textbf{Stage 1} (left panel of the figure), we train the Fast Dynamic-Aware Policy and a variational autoencoder (VAE). In \textbf{Stage 2} (right panel of the figure), we apply training-free, per-step action corrections using the aforementioned Fast Dynamic-Aware Policy; the corrected actions are then decoded by the VAE decoder.
}
\label{fig:method}
\vspace{-10pt}
\end{figure*}

\subsection{Dynamic Manipulation}
In robot-object interactions, objects are often in motion, such as when handling items on an industrial conveyor belt or interacting with dynamic objects like a soccer ball~\cite{labiosa2025reinforcement}, a badminton shuttlecock~\cite{shi2025mv}, or a table tennis ball~\cite{su2025hitter} during sports activities.
However, there is currently no dedicated real or simulated dataset specifically designed to train robots for adapting to such dynamic scenarios.
To address dynamic challenges, some approaches~\cite{hansen2022temporal,salzmann2023real} integrate Model Predictive Control (MPC)~\cite{rawlings2020model} to achieve real-time performance. While effective in structured environments, these methods tend to generalize poorly when the motion sequences of the manipulated objects become more uncertain. 
Additionally, some methods~\cite{zhang2025catch,zhang2024graspxl,zhang2026instructionanchorsdissectingcausal} combine reinforcement learning to explore dynamic capabilities. However, these approaches do not fully leverage the action expert data derived from imitation learning, as discussed in Section~\ref{sec:BC}. 
In contrast, our approach injects dynamic correction information into the diffusion policy, which has been trained through imitation learning, in a training-free manner during the second-stage dynamic scene evaluation. This method eliminates the need for additional training of the action model and enhances the model’s ability to operate effectively in dynamic environments.

%% file: section/3_method.tex
\section{\textbf{Method}}
\label{sec:method}

\subsection{\textbf{Preliminaries}}

\textbf{Diffusion policy with action chunking.}
Let $\mathbf{A}_{t:t+H-1} \coloneqq [\mathbf{a}_t,\mathbf{a}_{t+1},\ldots,\mathbf{a}_{t+H-1}]$ denote a horizon-$H$ action chunk conditioned on the current observation $\mathbf{o}_t$. We instantiate the slow policy as a conditional diffusion model,$\pi_s(\mathbf{A}_{t:t+H-1}\mid \mathbf{o}_t)$, which transforms Gaussian noise into $\mathbf{A}_{t:t+H-1}$ via an observation-conditioned denoiser $\epsilon_{\theta_s}$. During dynamic evaluation we keep the parameters $\theta_s$ fixed (training-free), reusing the weights learned on a static dataset. While $\pi_s$ promotes temporal coherence across the chunk, it operates open-loop within the chunk and thus cannot promptly react to object motion in rapidly changing scenes (Please see Figure~\ref{fig:method} Stage 2). 

\textbf{Dynamics-aware policy.}
We maintain a history bank of the most recent $M$ observations (Figure~\ref{fig:method} Stage 1),
$\mathbf{O}_{t-M+1:t} \coloneqq [\mathbf{O}_{t-M+1},\ldots,\mathbf{O}_t].$
A fast policy extracts dynamics-aware, memory features from this history,
$\mathbf{F}_M = \pi_f(\mathbf{O}_{t-M+1:t}).$

\textbf{Joint policy for dynamic evaluation.}
At dynamic evaluation time, we fuse the slow and fast pathways through a joint policy
$\pi_c(\mathbf{A}'_{t:t+H-1}\mid \mathbf{o}_t,\mathbf{F}_M),$
where $\mathbf{A}'_{t:t+H-1}$ is the action chunk after online dynamic correction, jointly predicted from the current observation $\mathbf{o}_t$ and the features $\mathbf{F}_M$. In the following, Section~\ref{subsec:stage1} details the training of the fast dynamic aware policy $\pi_f$, and Section~\ref{subsec:stage2} explains how its features condition the diffusion policy to yield corrected action chunks in dynamic environments.

\subsection{\textbf{Stage 1: Fast Dynamic Aware Policy Training}}
\label{subsec:stage1}

This section details the Stage 1 training procedure of the Fast Dynamic-Aware Policy, as illustrated in Figure~\ref{fig:method} (left). We use a History Bank to store a sliding window of observations of size $M$, denoted as $\mathbf{O}_{t-M+1:t}$. The History Bank is then used in conjunction with the computed differential features for cross-attention, aiming to capture dynamic information, while utilizing frame-wise temporal attention to learn the sequential history (see Section~\ref{subsubsec:history bank}). Additionally, we compute a self-supervised loss using the differential features for the output of the fast dynamic aware policy, and train the model accordingly (see Section~\ref{subsubsec:self super}). To ensure that the Fast Dynamic Aware Policy training fully leverages expert data, we employ a lightweight Variational Autoencoder for action prediction training on pushT data. In this case, the output of the Fast Dynamic Aware Policy is jointly used with $\mathbf{o}_t$ to predict the action and compute the loss for backpropagation (see Section~\ref{subsubsec:vae}).

\subsubsection{\textbf{History Bank Memory Learning}}
\label{subsubsec:history bank}

We propose a Dynamic Feature Extractor, which is designed to capture both temporal dependencies and dynamic changes in the environment. The system relies on a sliding window of the most recent observations, stored in the \textbf{History Bank}, to extract dynamic features using a combination of convolutional layers and attention mechanisms.

Let $\mathbf{O}_{t-M+1:t} \coloneqq [\mathbf{O}_{t-M+1}, \mathbf{O}_{t-M+2}, \ldots, \mathbf{O}_t]$ represent the history bank containing the most recent $M$ observations. These observations are fed into the feature extractor to capture spatial and temporal dependencies. Initially, the input $\mathbf{O}_{t-M+1:t}$ is processed through a pre-trained ResNet18 backbone to extract spatial features. Let the extracted feature map be denoted by:
\begin{equation}
\mathbf{X}_{\text{spatial}} = \text{ResNet}(\mathbf{O}_{t-M+1:t}),
\end{equation}
where $\mathbf{X}_{\text{spatial}} \in \mathbb{R}^{M \times C \times H_f \times W_f}$; here $M$ is the number of frames, and $C, H_f, W_f$ denote the channel, height, and width of the feature maps, respectively. 

\subsubsection{\textbf{Differential Feature Computation}}
To capture the dynamic changes between consecutive frames, we compute the differential feature $\Delta \mathbf{X}_{t}$ as:
\begin{equation}
\Delta \mathbf{X}_t = \mathbf{X}_{t+1} - \mathbf{X}_t,
\end{equation}
where $\mathbf{X}_{t}$ represents the extracted feature map for frame $t$. We scale the differential by a learnable parameter $\alpha$:
\begin{equation}
\mathbf{D}_t = \alpha \cdot (\mathbf{X}_{t+1} - \mathbf{X}_t),
\end{equation}
where $\mathbf{D}_t \in \mathbb{R}^{(M-1) \times C' \times H_f \times W_f}$ represents the differential features for the sliding window. This operation allows the model to capture temporal dynamics between adjacent frames.

\subsubsection{\textbf{Temporal Attention}} 
The Temporal Attention mechanism is designed to capture dependencies across the time dimension in a sequence of frames. It enables the model to learn temporal relationships between frames, allowing it to focus on the most relevant time steps and capture long-range temporal dependencies. 

Given frame features $\mathbf{X}_{\mathrm{spatial}} \in \mathbb{R}^{M \times C' \times H_f \times W_f}$, the model computes, for each time step $t$, the \textbf{query} $(\mathbf{Q}_t)$, \textbf{key} $(\mathbf{K}_{t'})$, and \textbf{value} $(\mathbf{V}_{t'})$, obtained from $\mathbf{X}_{\mathrm{spatial}}$ via learned linear projections. The attention scores are then computed by taking the dot product between the query for time step $t$ and the key for time step $t'$, with the attention score $\text{Attn}_{t,t'}$ given by:
\begin{equation}
\text{Attn}_{t,t'} = \text{SoftMax}(\frac{\mathbf{Q}_t \cdot \mathbf{K}_{t'}^T}{\sqrt{D}}),
\end{equation}
where $D$ denotes the dimensionality of the query and key vectors; the score quantifies temporal similarity between time steps $t$ and $t'$ and is scaled by $1/\sqrt{D}$ to stabilize gradients during training. The attention scores are then normalized using the \textbf{softmax} function to convert them into a probability distribution. 

Finally, the attention scores are applied to the values $\mathbf{V}_{t'}$, producing the attended output for time step $t$. The output is the weighted sum of the values across all time steps, allowing the model to focus on the most important frames. The attended features for time step $t$ are computed as:
\begin{equation}
\mathbf{X}_{\text{attended}, t} = \sum_{t'} \text{Attn}_{t,t'} \cdot \mathbf{V}_{t'}.
\end{equation}
The final attended features are then passed through a linear projection to produce the output tensor $\mathbf{X}_{\text{temporal}}$. 

The Temporal Attention mechanism effectively learns long-range temporal dependencies by focusing on relevant time steps, making it especially powerful for modeling dynamic environments where the relationships between past and future frames are crucial for accurate predictions.

\subsubsection{\textbf{Fusion Cross-Attention}}
To relate dynamic features to the observation history, we apply cross-attention to fuse the differential feature $\mathbf{D}_t$ with temporal context from the history bank $\mathbf{X}_{\text{temporal}}$. 
Given queries $\mathbf{Q}_t$ from $\mathbf{X}_{\text{temporal}}$ and keys/values $\mathbf{K}_t,\mathbf{V}_t$ from $\mathbf{D}_t$, the attention output is
\begin{equation}
\operatorname{Attn}(\mathbf{Q}_t,\mathbf{K}_t,\mathbf{V}_t)
= \operatorname{softmax}\!\left(\frac{\mathbf{Q}_t \mathbf{K}_t^{\top}}{\sqrt{d_k}}\right)\mathbf{V}_t,
\end{equation}
where $d_k$ is the key dimensionality. 
This cross-attention aligns temporal context with differential cues, enabling the model to attend to dynamic changes over time. 
Let $\mathbf{F}_M$ denote the fused representation that summarizes both historical memory and dynamic variation. 

\subsubsection{\textbf{Self-Supervised with Differential}}
\label{subsubsec:self super}
In this section, we present a self-supervised learning scheme for the dynamic feature extractor that leverages frame-to-frame differentials to model temporal change, removing the need for manual labels.

During training, at step $M$ the extractor produces predicted dynamic features $\mathbf{F}_M$, while the history bank provides differential targets $\mathbf{D}_{M-1}$ (Sec.~\ref{subsubsec:history bank}). The model is conditioned on preceding frames, and $\mathbf{D}_{M-1}$ serves as supervision.

To align predictions with observed changes, we minimize the KL divergence between normalized features:
\begin{equation}
\mathcal{L}_{\mathrm{diff}}
= \sum_{t=1}^{T} \mathrm{KL}\!\left(\operatorname{softmax}(\mathbf{F}_M^{(t)})\,\|\,\operatorname{softmax}(\mathbf{D}_{M-1}^{(t)})\right).
\end{equation}
This objective encourages the representation to capture temporal dynamics without manual annotations.

\subsubsection{\textbf{Variational Autoencoder}}
\label{subsubsec:vae}

We utilize a modified \textbf{Variational Autoencoder (VAE)} to predict future actions. The model consists of an encoder that processes the action sequence and a decoder that generates the predicted future actions conditioned on dynamic temporal features. This architecture allows the model to learn a compact latent representation that captures the essential temporal dynamics between the current and future actions.

\textbf{Encoder and Latent Space Representation.}
The \textbf{encoder} network takes an action sequence $\mathbf{A}_t$ as input, and outputs the \textbf{mean} ($\mu$) and \textbf{log-variance} ($\log(\sigma^2)$) of a Gaussian distribution, which parameterizes the approximate posterior distribution $q(\mathbf{z} \mid \mathbf{A}_t)$ over the latent variable $\mathbf{z}$:
\begin{equation}
q(\mathbf{z} \mid \mathbf{A}_t) = \mathcal{N}(\mu, \sigma^2).
\end{equation}
This distribution is used to capture the underlying factors of variation in the action sequence. The latent vector $\mathbf{z}$ is then sampled from this distribution using the reparameterization trick to allow for backpropagation through the stochastic sampling process:
\begin{equation}
\mathbf{z} = \mu + \sigma \cdot \epsilon,
\end{equation}
where $\epsilon \sim \mathcal{N}(0, I)$ is random noise, and $\sigma = \exp\left(\frac{1}{2} \log(\sigma^2)\right)$ is the standard deviation derived from the log-variance.

\textbf{Decoder and Action Prediction.}
The decoder generates the predicted future actions $\hat{\mathbf{A}}_t$, conditioned on both the latent vector $\mathbf{z}$ and the dynamic features $\mathbf{F}_M$ of the action sequence. These dynamic features $\mathbf{F}_M$ capture the temporal dependencies of the environment, which are used as additional context for action prediction. The decoder reconstructs the action sequence using a \textbf{recurrent neural network (RNN)} that takes in the latent vector and the temporal context to output the predicted action sequence:
\begin{equation}
p(\mathbf{A}_t \mid \mathbf{z}, \mathbf{F}_M) = \mathcal{N}(\hat{\mathbf{A}}_t, \sigma_{\text{decoder}}^2).
\end{equation}
Here, $\hat{\mathbf{A}}_t$ is the predicted action, and $\sigma_{\text{decoder}}$ is the standard deviation of the predicted distribution.

\subsubsection{\textbf{Loss}}
\label{subsubsec:loss}

The training objective of the VAE is a combination of the \textbf{reconstruction loss} and the \textbf{KL divergence} regularization. The reconstruction loss is computed as:
\begin{equation}
\mathcal{L}_{\text{recon}} = \text{MSELoss}(\hat{\mathbf{A}}_t, \mathbf{A}_t).
\end{equation}
The KL divergence is computed as:
\begin{equation}
\mathcal{L}_{\text{KL}} = -\frac{1}{2} \sum_{i=1}^{d} \left( 1 + \log(\sigma_i^2) - \mu_i^2 - \sigma_i^2 \right),
\end{equation}
Where $\mu_i$ and $\sigma_i$ are the mean and standard deviation of the posterior for the $i$-th latent dimension. The differential loss $\mathcal{L}_{\text{diff}}$ is computed using the KL divergence as mentioned above.  Thus, the total loss function is the weighted sum of the reconstruction loss, the KL divergence, and the differential loss:
\begin{equation}
\mathcal{L}_{\text{total}} = \mathcal{L}_{\text{recon}} + \lambda_{\text{KL}} \mathcal{L}_{\text{KL}} + \lambda_{\text{diff}} \mathcal{L}_{\text{diff}},
\end{equation}
where $\lambda_{\text{KL}}$ and $\lambda_{\text{diff}}$ are hyperparameters controlling the relative importance of each term. The forward pass involves the action sequence $\mathbf{A}_t$ and the temporal conditional features $\mathbf{F}_M$, and the VAE model outputs the reconstructed actions $\hat{\mathbf{A}}_t$, along with the mean $\mu$ and log-variance $\log(\sigma^2)$ from the encoder. The total loss is then backpropagated to update the model parameters.

\subsection{\textbf{Stage 2: Dynamic Injection for Training-Free Diffusion Policy}}
\label{subsec:stage2}

During dynamic evaluation, we maintain a sliding window of length $M$ over the observations,
$\mathbf{O}_{t-M+1:t} = [\mathbf{o}_{t-M+1}, \ldots, \mathbf{o}_t]$.
We then compute dynamics-aware features online using the Stage-1 Fast Dynamic-Aware extractor $\pi_f$:
\begin{equation}
\mathbf{F}_t = \pi_f\!\left(\mathbf{O}_{t-M+1:t}\right) \in \mathbb{R}^{d_f}.
\end{equation}
A frozen slow diffusion policy produces a horizon-$H$ open-loop action chunk conditioned on the current observation,
\begin{equation}
\mathbf{A}_{t:t+H-1}\sim \pi_s(\,\cdot\,\mid \mathbf{o}_t),\quad 
\mathbf{A}_{t:t+H-1}=[\mathbf{a}_t,\ldots,\mathbf{a}_{t+H-1}].
\end{equation}

For feature injection, we first apply elementwise normalization using known bounds $\mathbf{a}_{\min}, \mathbf{a}_{\max}$:
\begin{equation}
\tilde{\mathbf{A}}_{t:t+H-1}
= 2\,\frac{\mathbf{A}_{t:t+H-1}-\mathbf{a}_{\min}}{\mathbf{a}_{\max}-\mathbf{a}_{\min}} - \mathbf{1},
\end{equation}
where all operations are elementwise and $\mathbf{1}$ denotes an all-ones tensor matching the shape of $\mathbf{A}_{t:t+H-1}$. 
We then encode this chunk into a latent vector with the frozen VAE encoder $E$ trained in Stage~1:
\begin{equation}
\mathbf{z}_t = E\!\left(\tilde{\mathbf{A}}_{t:t+H-1}\right)\in\mathbb{R}^{d_z}.
\end{equation}

Within each chunk, execution proceeds in a closed loop. 
For each step $s \in \{0,\ldots,H-1\}$, we refresh the history window and recompute features:
\begin{equation}
\mathbf{F}_{t+s} = \pi_f\!\left(\mathbf{O}_{t+s-M+1:t+s}\right).
\end{equation}
We then decode a dynamically corrected action with the frozen VAE decoder $D$ trained in Stage~1, conditioned on a step embedding $\mathbf{e}_s$:
\begin{equation}
\hat{\mathbf{a}}_{t+s} = D\!\left(\mathbf{z}_t,\, \mathbf{F}_{t+s},\, \mathbf{e}_s\right).
\end{equation}
Aggregating over steps yields the closed-loop chunk:
\begin{equation}
\hat{\mathbf{A}}'_{t:t+H-1}
= \left[\, D(\mathbf{z}_t,\mathbf{F}_{t},\mathbf{e}_0),\ \ldots,\ D(\mathbf{z}_t,\mathbf{F}_{t+H-1},\mathbf{e}_{H-1}) \,\right].
\end{equation}

Equivalently, this induces a deterministic joint closed-loop policy
\begin{equation}
\begin{split}
\pi_c(\hat{\mathbf{a}}_{t+s} \mid \mathbf{o}_t, \mathbf{O}_{t+s-M+1:t+s})
= & \delta( \hat{\mathbf{a}}_{t+s} - D(E(\mathrm{norm}(\pi_s(\mathbf{o}_t))), \\
  & \ \pi_f(\mathbf{O}_{t+s-M+1:t+s}), \mathbf{e}_s))
\end{split}
\end{equation}
where $\mathrm{norm}(\cdot)$ denotes the linear normalization above and $\delta(\cdot)$ is a Dirac delta indicating a deterministic mapping. After $s=H-1$, we replan by resampling a new chunk from $\pi_s$ with the latest observation (receding-horizon), while keeping all modules frozen; thus, the entire Stage-2 procedure is \emph{training-free}. The continual injection of $\mathbf{F}_{t+s}$ provides fine-grained online corrections to the open-loop diffusion chunk, improving responsiveness and robustness in rapidly changing scenes.

%% file: section/4_experiment.tex
\section{Experiment}
\label{sec:exp}
In this section, we evaluate the proposed algorithm on the PushT task in dynamic scenarios.
First, under static settings, we evaluate the effect of the method on task success rate and compare it with the standard Diffusion Policy inference strategy.
Then, in dynamic scenarios, we further investigate the robustness and generalization of the algorithm, and conduct ablation studies to validate the effectiveness of each design module.
Finally, we measure the inference latency of the algorithm, demonstrate its lightweight nature, and show that it can support high-frequency closed-loop control and real-time deployment.

\subsection{Experimental Settings}
For the PushT task, we employ a Diffusion Policy trained from human demonstrations as the basic control strategy.
During inference, we set the batch size to $N=50$, meaning that 50 initial poses are randomly generated in the simulation environment and kept fixed throughout the experiments.

For each initial condition, we conduct rollouts in the environment and evaluate the performance of each method in terms of task success rate and inference latency. The simulation terminates either when the maximum number of steps $T_{\text{max}} = 300$ is reached, or earlier if the overlap ratio $\sigma$ between the object and the target position exceeds 95\%. The simulation environment operates at 10 Hz. The model predicts the next 16 actions in a single inference step and executes the first 8 actions in open-loop mode.

\textbf{Baselines}: We consider three representative existing inference methods as baselines for comparison:
\begin{itemize}[left=0pt] 
  \item Original Open-loop(H=8): Execute an entire action chunk at each inference step.
  \item Original Closed-loop(H=1): Execute only the most recent action at each inference step.
  \item Temporal Ensemble~\cite{zhao2023learning}(H=1): At each overlapping step, we average the new prediction $a$ with the previous prediction $\hat a$ to produce smoother action chunks: $a_t = \lambda a_t + (1 - \lambda) \hat a_t$, with $\lambda$ set to 0.5.
\end{itemize}
H denotes that the Diffusion Policy model is inferred every $H$ time steps.

\textbf{Perturbations:}To systematically evaluate the robustness and generalization of the proposed method under dynamic conditions, we introduce two types of perturbations in simulation:
\begin{itemize}[left=0pt]
  \item Constant-direction Perturbations: During each execution step, an offset of fixed magnitude and direction is applied to the object to emulate a sustained external force.
  \item Random-direction Perturbations: During each execution step, an offset with fixed magnitude and randomly varying direction is applied, where the direction is resampled every $N=50$ steps.
\end{itemize}
\begin{figure}[htbp] 
  \centering
  \includegraphics[width=0.9\columnwidth]{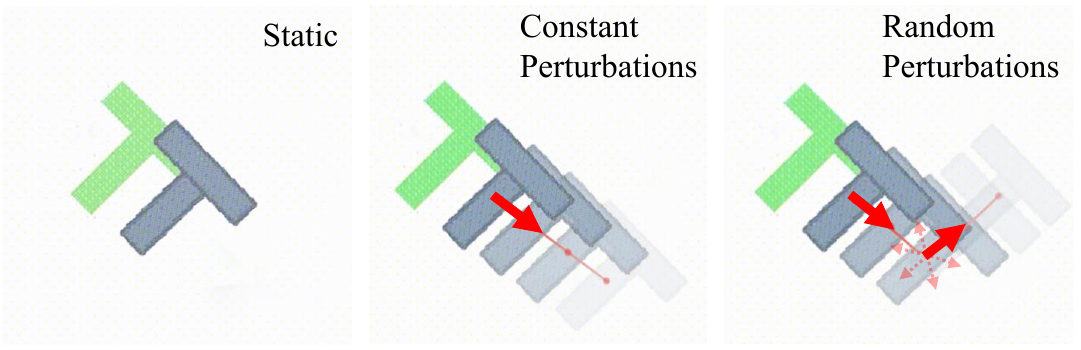}
  \caption{Different types of perturbations are considered, where the random perturbation updates its direction at certain time steps.}
  \label{fig:perturbations}
  \vspace{-10pt}
\end{figure}

\textbf{Dataset}: This study employs a publicly available human demonstration dataset consisting of 200 trajectories, where the blocks are free from external Perturbations.

\textbf{Model Training}: Our method was trained from scratch on a single NVIDIA A800 80GB GPU. For comparison, as a baseline, the Diffusion Policy is trained with default parameters for 500 epochs to ensure convergence.

\textbf{Inference Latency Measurement}: To ensure a fair evaluation of inference efficiency, we measure all inference latencies on a single NVIDIA RTX 4090 GPU 24GB under identical hardware and software configurations.
\subsection{Quantitative Analysis}
Table~\ref{tab:sr} presents a comparison of task success rates between the proposed method and several baselines in both static scenarios and dynamic scenarios under varying Perturbations. 
\begin{table}[htbp]
  \caption{Success rates across three levels of perturbation intensity show that the proposed DCDP method outperforms all baseline methods.}
  \centering
  \label{tab:sr}
  \normalsize
  \begin{tabular*}{\columnwidth}{@{\extracolsep{\fill}} c|ccc} 
    \toprule
    Methods & Static & \makecell{Constant \\ Perturbations} & \makecell{Random \\ Perturbations}  \\
    \midrule
    Open-Loop & 88.4 & 58.2 & 52.8  \\
    Close-Loop & 84.6 & 76.1 & 61.6  \\
    \makecell{Temporal \\ Ensemble} & 81.0 & 65.8 & 57.3  \\
    \textbf{DCDP} & \textbf{92.5} & \textbf{77.6} & \textbf{71.9}  \\
    \bottomrule
  \end{tabular*}
  \vspace{-10pt}
\end{table}

Table~\ref{tab:delay} shows the average single-step latency for each method, highlighting the real-time and computational improvements of our method.
\begin{table}[htbp]
\centering
\caption{Comparison of per-step inference latency. DCDP adds only ~5\% overhead in closed-loop execution, substantially lower than other closed-loop methods.}
\normalsize 
\setlength{\tabcolsep}{3pt} 
\begin{tabular*}{\columnwidth}{@{\extracolsep{\fill}} ccccc}
\toprule
Methods & OL(H=8) & CL(H=1) & TE(H=1) & DCDP(H=8) \\
\midrule
Delay (ms) & 7.05 & 53.60 & 53.74 & \textbf{7.39} \\
\bottomrule
\end{tabular*}
\vspace{-10pt}
\label{tab:delay}
\end{table}

We observe that the original closed-loop policy outperforms the open-loop policy in terms of task success rate under dynamic scenarios, whereas it exhibits a marked performance drop in static scenarios.
We speculate that this is because, as a diffusion model, Diffusion Policy replans the entire action sequence during each inference step, which results in discontinuity between consecutive actions.
This may impair its ability to reproduce long-term coordination in human demonstrations, thus reducing the task success rate.
In contrast, DCDP operates at a lower replanning frequency, exploiting the long-term planning ability of the diffusion policy while integrating recent observations for rapid closed-loop control, resulting in notable improvements in both static and dynamic tasks.
\begin{figure}[htbp]
  \centering
  \includegraphics[width=0.95\columnwidth]{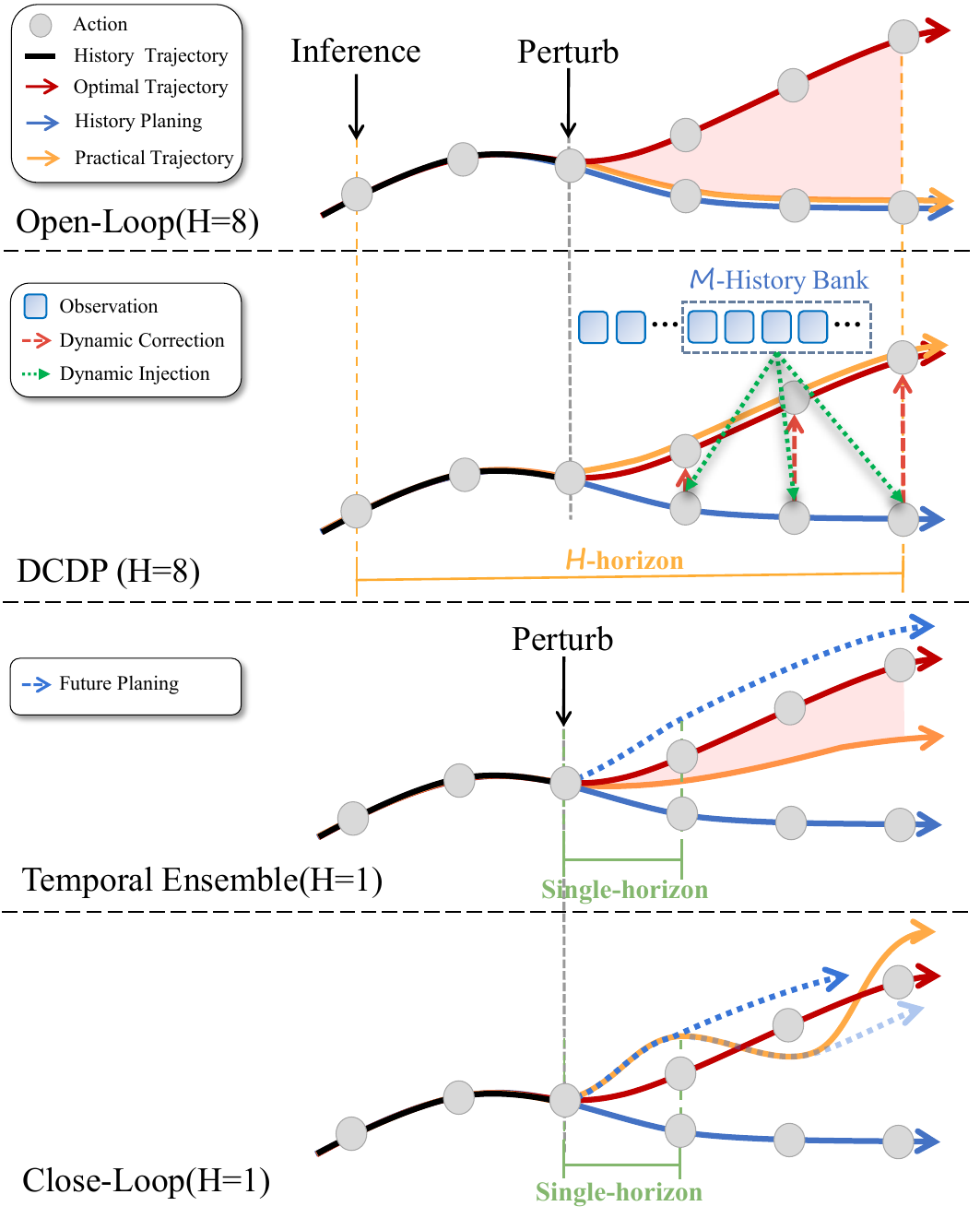}
  \caption{Visualization of how various inference strategies respond to perturbations.}
  \vspace{-12pt}
  \label{fig:diff-methods}
\end{figure}

In addition, the inference latency data show that our lightweight model has clear advantages in real-time performance, with lower latency than the direct closed-loop strategy and supporting high-frequency closed-loop control
 
\subsection{Ablation Study}
To assess the necessity and effectiveness of each module in the proposed architecture, we conduct ablation studies to analyze their relative contributions to model performance.

Specifically, we conducted ablation studies on the self-supervised dynamic feature extraction module. Key components were sequentially removed, and Stage 1 was retrained after each removal. Training parameters and evaluation metrics were kept the same as in the original experiment.
\begin{table}[htbp]
  \centering
  \caption{Task success rates after ablating individual modules, demonstrating that each module in DCDP is effective.}
\normalsize
\begin{tabularx}{\columnwidth}{XXX|XX}
\toprule
TA &SSD &DCA& 
SR$_{\text{static}}$ &
SR$_{\text{dynamic}}$ \\
\midrule
 &  &  &88.40 & 58.20 \\
 & \checkmark & \checkmark &92.05 & 73.59 \\
 \checkmark&  & \checkmark& 92.31 & 73.50 \\
\checkmark & \checkmark & & 93.36 & 68.23 \\
\checkmark & \checkmark & \checkmark &  \textbf{92.50} & \textbf{77.60} \\
\bottomrule
\end{tabularx}
\label{tab:ablation}
\end{table}

Table~\ref{tab:ablation} presents the results of the ablation experiments.\textbf{TA} refers to temporal attention, \textbf{SSD} refers to self-supervised diff loss, \textbf{DCA} refers to diff cross attention, and \textbf{SR$_{\text{static}}$}, \textbf{SR$_{\text{dynamic}}$}, denotes success rate (\%). We selected static scenarios and constant-direction perturbations as evaluation metrics.

The results demonstrate that all proposed modules make positive contributions to system performance.

\subsection{Real World Application}
To validate the effectiveness of the proposed algorithm in real-world scenarios, we select two representative manipulation tasks for evaluation:
\begin{itemize}[left=0pt]
  \item Pick-and-place of a moving cup, corresponding to the constant-direction perturbation scenario, which examines the algorithm’s robustness under continuous perturbations;
  \item Pouring liquid into a moving cup, corresponding to the random-direction perturbation scenario, which evaluates the algorithm’s adaptability to uncertain perturbations.
\end{itemize}

We employed the UMI~\cite{chi2024universal} gripper to perform the tasks and utilized the publicly available FastUMI dataset, reproducing its scenarios to construct a real-world evaluation platform.
\begin{figure}[t]
  \centering
  \includegraphics[width=0.9\columnwidth]{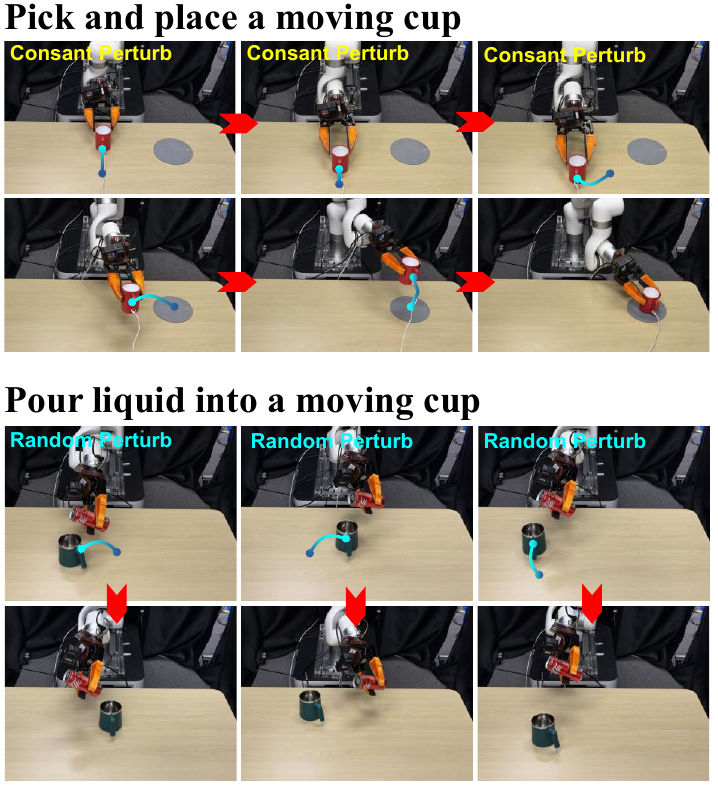}
  \caption{The two tasks comprise two types of perturbations: constant-direction and random-direction. These perturbations were applied exclusively to the components highlighted in the figure.}
  \label{fig:realworld}
  \vspace{-15pt}
\end{figure}

In the real-world task tests, we adopted the PushT dynamic task evaluation design and implemented constant- and random perturbations. Constant-direction perturbations were applied in the "picking and placing a moving cup" task, in which the cup moved along a single direction prior to grasping. Random perturbations were applied in the "pouring liquid into a moving cup" task, in which the cup’s movement direction was random, requiring the robotic arm to pour liquid accurately into it.

The results indicate that, compared with the original inference method, DCDP exhibits significantly enhanced adaptability in dynamic environments and can handle a range of dynamic scenarios.

%% file: section/5_conclusion.tex
\section{Conclusion}
\label{sec:conclusion}

We propose DCDP, a training-free closed-loop action-chunk framework that injects high-frequency dynamic features into a pretrained diffusion policy for real-time correction. Key elements include a self-supervised dynamic feature encoder, cross/temporal attention for temporal awareness, and an asymmetric action encoder/decoder that decodes frozen diffusion chunks with updated context. 
Since correction happens at inference without retraining, DCDP is a lightweight, plug-and-play balance between long-horizon planning and responsive control. Current limits include evaluation on a single simulated task and small data; future work should test diverse tasks, real hardware, and adaptive scheduling with multi-modal dynamics.